\documentclass{article}
\usepackage{spconf,amsmath,graphicx,hyperref}

\usepackage{multirow}
\usepackage{multicol}
\usepackage{dblfloatfix}
\usepackage{enumitem}
\usepackage{lipsum} % for dummy text
\usepackage{wrapfig} % for wrafpping text around figures
\usepackage{algpseudocode}
\usepackage{algorithm}
\usepackage{algpseudocode}
\usepackage{amsmath,amssymb}
\usepackage{xcolor}         % colors
\usepackage{color, colortbl}
\usepackage{makecell}
\usepackage{pifont}
\usepackage{cleveref}
\usepackage{tabularx}
\usepackage{pifont}
\usepackage{tabularx}
\usepackage{tcolorbox}
\usepackage[subtle]{savetrees}
\definecolor{mygreen}{RGB}{17, 128, 41}
\definecolor{myred}{RGB}{185, 0, 27}
\newcommand{\methodname}{WAVECLIP}

\usepackage{mathtools,bm}

\title{WAVECLIP: Wavelet Tokenization for Adaptive-Resolution CLIP}

\name{%
  Moshe Kimhi$^{1,2}$\sthanks{Equal contribution} \qquad
  Erez Koifman$^{1}$ \footnotemark[1] \qquad
  Ehud Rivlin$^{1}$ \qquad
  Eli Schwartz$^{2}$\sthanks{Equal supervision} \qquad
  Chaim Baskin$^{3}$\footnotemark[2]
}
\address{$^{1}$ Technion -- Israel Institute of Technology \quad $^{2}$ IBM Research \\
 $^{3}$ Ben-Gurion University of the Negev} %\quad$^{3}$ Google
\address{$^{1}$ Technion -- Israel Institute of Technology \quad $^{2}$ IBM Research \\
$^{3}$ Ben-Gurion University of the Negev} % \quad $^{3}$ Google
\begin{document}
%\ninept
%
\maketitle
\begin{abstract}
We introduce WAVECLIP, a single, unified model for adaptive-resolution inference in CLIP, enabled by wavelet-based tokenization. WAVECLIP replaces standard patch embeddings with a multi-level wavelet decomposition, enabling the model to process images coarse-to-fine while naturally supporting multiple resolutions within the same model. At inference time, the model begins with low-resolution tokens and refines only when needed, using key–value caching and causal cross-level attention to reuse computation, effectively introducing to the model only new information when needed.
We evaluate WAVECLIP in zero-shot classification, demonstrating that a simple confidence-based gating mechanism enables adaptive early exits. This allows users to dynamically choose a compute–accuracy trade-off using a single deployed model. Our approach requires only lightweight distillation from a frozen CLIP teacher and achieves competitive accuracy with significant computational savings.
\end{abstract}
\section{Introduction}
\label{sec:intro}
Contrastive language-image pretraining (CLIP) enables strong zero-shot transfer with a frozen text tower and a Vision Transformer (ViT) image encoder \cite{radford2021clip,cherti2023reproducible,zhai2023siglip}. However, its inference cost scales quadratically with the number of image tokens, leading to very high GFLOPs.

While much of the prior work on efficient CLIP reduces or redesigns the \emph{vision tower} \cite{wu2023tinyclip,vasu2024mobileclip,faghri2025mobileclip2} or modifies the training objective \cite{zhai2023siglip,flip,tang2025tulip}, these approaches—though effective for accuracy or efficiency—typically require heavy pre-training on web-scale datasets (e.g., 2-5B examples). Moreover, each model is trained independently, yielding separate models for different scenarios. This makes it hard to dynamically adjust the trade-off at inference time and complicates deployment in systems that need adaptive behavior.

\noindent We introduce \textbf{\methodname}, a training-light, drop-in alternative to patch tokenization that turns a multi-level discrete wavelet transform (DWT) into a \emph{progressive, cacheable} inference schedule for CLIP. \methodname{} processes coarse tokens first (LL band) and exits early when confident; otherwise, it \emph{appends} level-wise detail tokens (LH/HL/HH). A block-causal cross-level attention mask allows tokens at level~$\ell$ to attend to tokens from levels $\le\ell$, so key/value states are reused and prior computation is never repeated. A short distillation from a frozen CLIP teacher preserves zero-shot alignment, avoiding full image-text pretraining.

\textbf{Our contributions:} (i) A wavelet tokenizer + cross-level attention mask that enables early exits with KV reuse in CLIP, no ViT/text-tower changes. (ii) A light distillation procedure that maintains zero-shot alignment. (iii) Accuracy/compute trade-offs via simple confidence gating, that is controllable at inference time.

\begin{figure}[t]
  \centering
  \includegraphics[width=\columnwidth]{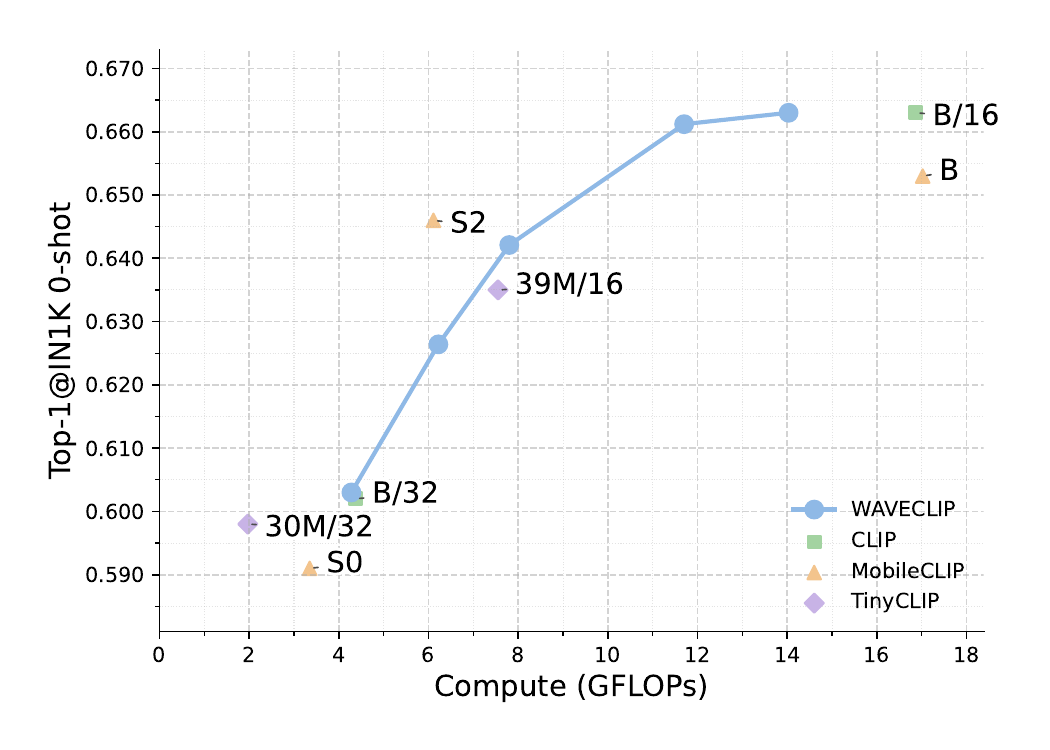}
  \vspace{-12mm}
\caption{\textbf{Accuracy–compute frontier on ImageNet-1k (zero-shot).}
\textbf{WAVECLIP} spans a smooth trade-off curve by adjusting threshold at inference, while prior methods appear as fixed points—each requiring a separate model for a single compute budget.}
  \label{fig:acc_gflops}
\end{figure}

\begin{figure*}[t]
    \centering
\includegraphics[width=.9\linewidth]{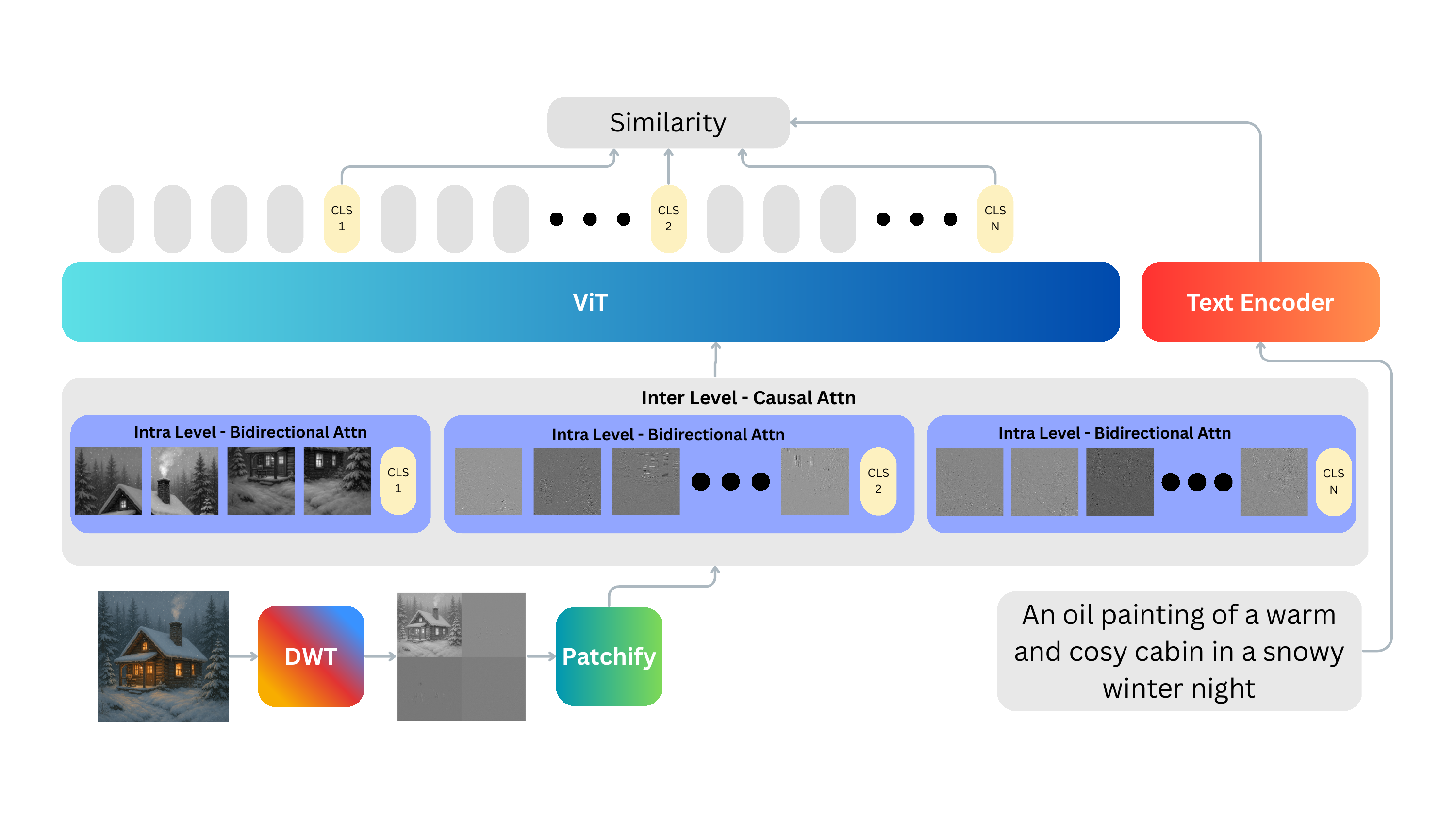}
    \vspace{-12mm}
    \caption{\textbf{WAVECLIP overview.} A discrete wavelet transform (DWT) decomposes the image and, after patchify, yields multiscale token groups. Inference proceeds coarse-to-fine in a single Transformer: each group uses bidirectional attention internally, while a causal cross-level mask lets later groups attend to earlier tokens only.}
    \label{fig:arch}
\end{figure*}

\section{Related work}
\label{sec:related}
In CLIP-specific efficiency, most prior work shrinks or redesigns the \emph{vision tower} itself:
TinyCLIP distills large CLIP models into compact students \cite{wu2023tinyclip};
MobileCLIP trains lightweight backbones with multimodal reinforcement \cite{vasu2024mobileclip};
and MobileCLIP2 further strengthens distillation/captioning for improved zero-shot at similar runtimes \cite{faghri2025mobileclip2}.
These approaches trade parameters and capacity for speed. \emph{In contrast, we keep the standard CLIP ViT and text tower intact} \cite{dosovitskiy2021vit,radford2021clip,cherti2023reproducible,zhai2023siglip} and target \emph{how much} visual evidence is processed per image.

Early-exit methods dynamically adjust compute based on input difficulty, by forecasting predictions from earlier layers of the model \cite{teerapittayanon2016branchynet,kaya2019shallowdeep}; this differs from our approach, as deciding to process fewer (coarser) tokens directly reduces GFLOPs when the model architecture stays intact, enabling robust deployment and KV-cache reuse.

\textit{Multiscale tokenization.} Wavelet decompositions provide a principled, dyadic multiresolution representation \cite{mallat1989theory,daubechies1992tenlectures} and have recently been explored as drop-in tokenizers for ViTs \cite{zhu2024wavelettokenizer}. We leverage this structure not only to replace patchify, but to enable \emph{progressive, cacheable} inference: start from the coarse LL band and \emph{only} add LH/HL/HH detail when the gate deems it necessary. A block-causal cross-level attention mask and key-value (KV) caching let finer levels reuse keys/values from coarser levels, avoiding recomputation while preserving the standard CLIP architecture. 

\vspace{-5mm}
\section{Method}
\label{sec:method}

Given an image $x\in\mathbb{R}^{H\times W\times 3}$ and a ViT image encoder with $B$ transformer blocks, a standard patch embedder with patch size $P$ yields $N=\frac{HW}{P^2}$ spatial tokens (plus a \texttt{[CLS]} token). Self-attention incurs $O(N^2d)$ cost per block, so reducing $N$ directly lowers GFLOPs. In CLIP-style zero-shot inference, an image embedding $v\in\mathbb{R}^d$ is compared to a bank of text embeddings $\{t_m\}_{m=1}^{M}$ via cosine similarity.

\begin{figure}[t]
  \centering
  \includegraphics[width=\columnwidth]{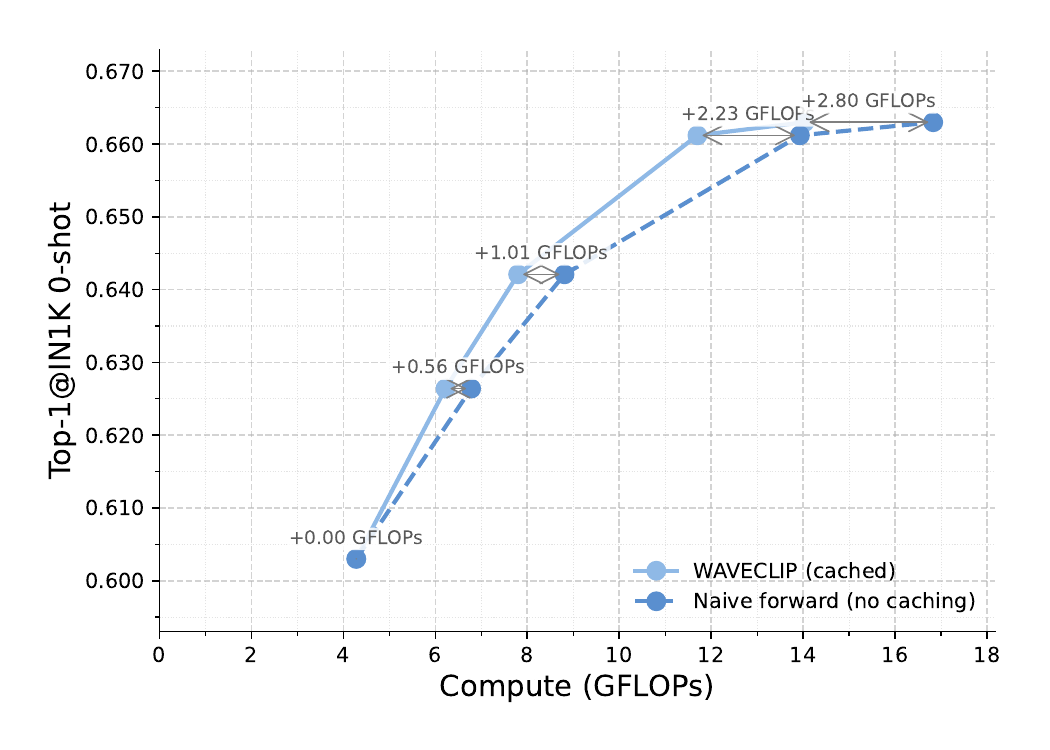}
  \vspace{-10mm}
  \caption{Comparing the computational cost of using KV cache vs no KV cache reuse.}
  \label{fig:cached_vs_naive}
\end{figure}

\subsection{Wavelet decomposition as structured tokenization}
\label{sec:wavelet}
Instead of patchify, we first convert $x$ from RGB to YCbCr and apply an $L$-level 2D discrete wavelet transform (DWT) \emph{independently to each channel}, producing channel-wise subbands
\[
\{(\mathrm{LL}^{(\ell,c)},\mathrm{LH}^{(\ell,c)},\mathrm{HL}^{(\ell,c)},\mathrm{HH}^{(\ell,c)})\}_{\ell=1}^{L}, c\in\{Y,\mathrm{Cb},\mathrm{Cr}\},
\]
with dyadic scaling \cite{mallat1989theory,daubechies1992tenlectures}; $\mathrm{LL}^{(L,c)}$ is the coarsest approximation for channel $c$, while $\mathrm{LH}/\mathrm{HL}/\mathrm{HH}$ add band-pass detail at progressively finer scales. We tokenize each subband by non-overlapping $P\times P$ patches and then \emph{merge} tokens from all channels, ordering them coarse-to-fine across levels (and channels), following evidence that wavelet tokenizers can replace patchify in ViTs \cite{zhu2024wavelettokenizer}. Unlike \cite{zhu2024wavelettokenizer}, which concatenates DWT subbands along the channel dimension to preserve the spatial grid and thus keep the token count $N$, we explicitly \emph{reduce} the number of tokens by processing coarse (LL) tokens first and appending LH/HL/HH detail only when needed, directly lowering the quadratic attention term and GFLOPs. Denote the cumulative token set after $\ell$ refinement steps by

$$
\mathcal{X}^{[0]}=\mathrm{patchify}(\mathrm{LL}^{(L)})
$$
$$
\mathcal{X}^{[\ell]}=\mathcal{X}^{[s-1]}\cup\mathrm{patchify}\!\big(\mathcal{D}^{(L-s+1)}\big),\; \ell=1,\dots,L
$$

where $\mathcal{D}^{(\ell)}=\{\mathrm{LH}^{(\ell)},\mathrm{HL}^{(\ell)},\mathrm{HH}^{(\ell)}\}$. Thus $\mathcal{X}^{[L]}$ equals the full-resolution token set. This design lets us begin with very few coarse tokens and add detail only when needed, directly controlling $N$.

\noindent \textbf{Token counts:}

At level $\ell$, spatial size  $\frac{H}{2^\ell}\times\frac{W}{2^\ell}$, yielding $\frac{HW}{P^2 4^\ell}$ tokens. Hence
\[
N_{\text{coarse}}=\frac{HW}{P^2 4^L}+1,\qquad
N^{[s]}=\frac{HW}{P^2}\!\left(\frac{1}{4^L}+\sum_{\ell=L-s+1}^{L}\frac{3}{4^\ell}\right)+s,
\]

where $+s$ accounts for \texttt{[CLS]}.

We train the model with the new 
DWT tokenizer and new \texttt{[CLS]} tokens to align with the original CLIP embedding space using a pretrained CLIP teacher.
Given a student embedding $v^{[\ell]}$ for the $\ell$'s \texttt{[CLS]} token, and teacher embedding $v_T$, we minimize the cosine distance:

\[
\mathcal{L}_{\text{distill}} = \sum_{\ell=1}^{L} 1 - \frac{\langle v^{[\ell]},\,v_T\rangle}{\|v^{[\ell]}\|\|v_T\|}
\]

% We run a short teacher-only distillation to match full-resolution image embeddings; no text supervision is required.

\subsection{Causal cross-level attention with KV reuse}
\label{sec:causal}
Let $f_\theta$ denote the ViT image encoder. We enforce \emph{level causality} via a block-lower-triangular (by level) attention mask: tokens introduced at level $\ell$ may attend to tokens from level $\le \ell$, but not vice versa. Formally, with cached key/value states $\mathcal{KV}^{[\ell-1]}$ from previous levels,
\begin{equation}
\label{eq:visuals}
v^{[s]} = f_\theta\!\big(\mathcal{X}^{[s]};\,\mathcal{KV}^{[s-1]},\,M_{\le s}\big),    
\end{equation}

where $M_{\le s}$ is the cross-level causal mask. This allows \emph{incremental} computation: upgrading from $s{-}1$ to $s$ reuses $\mathcal{KV}^{[s-1]}$ and primarily pays for interactions involving the new tokens, avoiding recomputation on earlier tokens.

Fig.~\ref{fig:arch} illustrates the tokenization~\ref{sec:wavelet} and causality~\ref{sec:causal}, While Fig.~\ref{fig:cached_vs_naive} shows the additional compute requires for naively forwarding all image tokens again, vs our KV cache reuse, advocating for the benefits of using Wavelet for tokenization.

\subsection{Progressive inference for zero-shot classification}
At level $\ell$, we score classes with temperature-scaled cosine similarity
\begin{equation}
\label{eq:similar}
s_m^{[\ell]}=\langle \mathrm{norm}(v^{[\ell]}),\,\mathrm{norm}(t_m)\rangle,\quad
m=1,\dots,M,    
\end{equation}

%add kimhi2025robot
Similar to \cite{kimhi2024,10943749} we apply a confidence gate to decide whether to \emph{exit} or to \emph{refine} (i.e., append the next finer level). We use a margin gate,
\begin{equation}
\label{eq:thresh}
s_{(1)}^{[\ell]}-s_{(2)}^{[\ell]} \;\ge\; \theta_m \;\;\Rightarrow\;\; \text{exit at level } \ell,
\end{equation}
where $s_{(1)}^{[\ell]}$ and $s_{(2)}^{[\ell]}$ are the top-1 and top-2 scores. Margin gating is empirically more stable across datasets than a fixed top-1 probability threshold, reducing the need for per-dataset calibration. When the condition is not met, we move to level $\ell{+}1$ (if $\ell \leq L$), reuse $\mathcal{KV}$, and evaluate again. This yields a single-pass behavior for easy images and coarse-to-fine refinement for hard images.

\begin{figure}[t]
  \centering
  \includegraphics[width=\columnwidth]{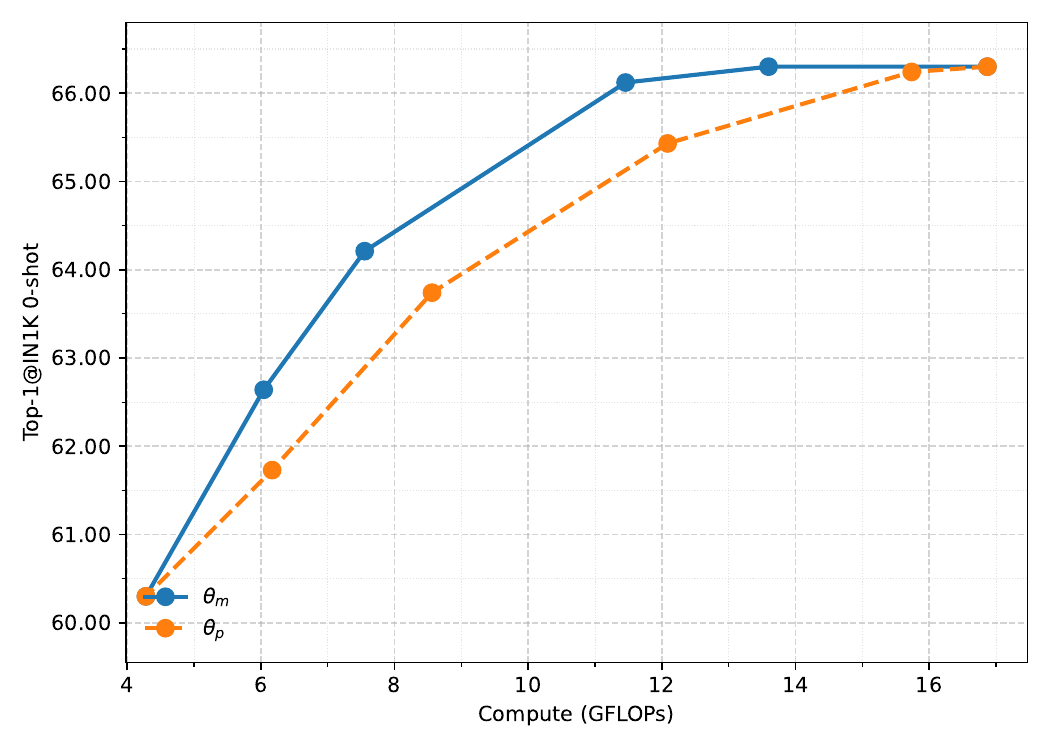}
  \vspace{-10mm}
  \caption{Margin ($\theta_m)$ vs.\ confidence ($\theta_p$) gating.}
  \label{tab:gating_strategies}
\end{figure}

We compare our strategy to simple thresholding the max prob. $s_{(1)}^{[\ell]} > \theta_p$. while simple, score thresholding must be \emph{carefully selected on a calibration set}, and, as shown in Fig.~\ref{tab:gating_strategies}, empirically harder to separate between samples that requires only coarse information.

\subsection{Complexity and control}
Because attention scales quadratically in token count, operating at smaller $N^{[s]}$ substantially reduces GFLOPs. The threshold $\theta_m$ acts as an \emph{inference-time knob} that trades accuracy for compute and can be adapted to constraints such as device budgets or low-battery modes, without modifying the trained model.
We choose $\theta_m$ to be relative to the number of classes in the dataset. e.g. $\theta_m = p \cdot N $  where $N$ is the number of classes in the dataset.

\begin{table}[h]
\centering
\caption{Token counts for $P{=}16$, $H{=}W{=}224$. $L$ is the number of DWT levels; $\ell$ is the number of refinement levels used. Counts include \texttt{[CLS]}; rows with $\ell{>}0$ also include one level marker per level.}
\label{tab:num_tokens}
\begin{tabular}{c|cccc}
\hline
Levels $L$ & $\ell{=}1$ & $\ell{=}2$ & $\ell{=}3$ & $\ell{=}4$ \\
\hline
$L{=}1$ & 197 & -- & -- & -- \\
$L{=}2$ & $50$ & $198$ & -- & -- \\
$L{=}3$ & $13$ & $51$ & $199$ & -- \\
$L{=}4$ & $4$ & $14$ & $52$ & $200$ \\
\hline
\end{tabular}
\end{table}

% 1. I would put attention mask side by side with the architecture (not above the lin. proj. part). Maybe add an arrow showing its relation the slef attention layer ( the colors are not enough to make this cear).

% 3. add to the visualization "RGB image" -> "wavelet image" -> wavelet patches"
% 4. maybe the attention mask will look better with the black block changed to white?

\begin{algorithm}
\caption{\methodname{} Progressive Inference}
\label{alg:waveclip}
\begin{algorithmic}[1]
\Require image $x$,Visual Encoder $f_\theta$ text embeddings $\{t_m\}$, threshold $\theta_m$
\State Compute $L$-level DWT: $\mathrm{LL}^{(L)}$, $\mathcal{D}^{(\ell)}$ for $\ell=1,\dots,L$
\State $s \gets 0$, $\mathcal{X}^{[0]} \gets \operatorname{patchify}(\mathrm{LL}^{(L)})$, $\mathrm{KV} \gets \varnothing$
\Repeat
\State Compute visual representation $v^{[s]}$  (Eq.~\ref{eq:visuals})
 % \State Run ViT on new tokens $\mathcal{X}^{[s]}\!\setminus\!\mathcal{X}^{[s-1]}$ with mask $M$; update $\mathrm{KV}$
  \State Compute scores $\mathbf{s}^{[s]}$ (Eq.~\ref{eq:similar})
  \If{Sufficient (Using Eq.~\ref{eq:thresh})}
    \State \Return prediction at level $\ell$
  \Else
    \State $\ell \gets \ell+1$
    \State $\mathcal{X}^{[\ell]} \gets \mathcal{X}^{[\ell-1]} \cup \operatorname{patchify}(\mathcal{D}^{(L-\ell+1)})$
  \EndIf
\Until{$\ell=L$}
\State \Return prediction with all image tokens at level $L$
\end{algorithmic}
\end{algorithm}

\section{Experiments}
\label{sec:experiments}

% \subsection{Setup}
% \label{sec:setup}

\textbf{Model} The teacher is a frozen CLIP ViT-B/16. \methodname{} replaces \emph{patchify} with an $L{=}2$-level discrete wavelet tokenizer while keeping the ViT image tower and the text tower unchanged.

\textbf{Progressive inference} At test time, tokens are processed coarse$\rightarrow$fine using a block-causal cross-level attention mask. Key–value states from coarser levels are cached and reused when adding finer tokens, so earlier computation is not repeated.

\textbf{Evaluation} We report zero-shot ImageNet-1k validation accuracy (Top-1) using standard CLIP prompts. Metrics include per-image GFLOPs (image tower) and average tokens processed. Figures~\ref{fig:acc_gflops} and \ref{fig:cached_vs_naive}, and Table~\ref{tab:main}, summarize the results, where for \methodname{}, \textbf{low,mid} and \textbf{high} refer to the margin threshold (yeilding ~26\%, 57\% and 74\% of samples require all tokens).  

All comparisons use 224$\times$224 input resolution. GFLOPs are reported for the \emph{image encoder only}; text embeddings are computed once per prompt set. For \methodname{} ``Tokens" column reports expected visual tokens. Baseline models were trained on datasets of up to $\approx$400M samples; for MobileCLIP and MobileCLIP2 we report accuracies from their longer training regimen (up to $\approx$1B seen samples). All accuracies are ImageNet-1k zero-shot with standard CLIP prompts.

\subsection{Accuracy vs Compute Trade-off}
\label{sec:tradeoff}
Fig.~\ref{fig:acc_gflops} shows that \methodname{} forms a smooth accuracy vs GFLOPs frontier using a \emph{single} deployed model with different gating thresholds.

\noindent\textbf{Operating points.} Using margin threshold $\theta_m$, it is possible to change the operating point for a deployed model. For example, with $p=0.03$, at \textbf{11.7 GFLOPs} \methodname{} attains \textbf{66.12\%}, yielding a \textbf{$\sim$30.6\%} compute reduction. At the high-accuracy end, where $p=0.5$ \methodname{}-\texttt{H} matches ViT-B/16 accuracy (66.3\%) at \mbox{14.03 GFLOPs}, a \textbf{$\sim$16.8\%} reduction over the baseline's \mbox{16.87 GFLOPs}. Tab.~\ref{tab:main} reports representative points, including the average tokens used by \methodname{} at each operating regime.
Unlike alternatives that train separate networks for different budgets, \methodname{} exposes a tunable accuracy-compute knob via early exits, simplifying deployment under dynamic constraints.

\begin{table}[t]
\centering
\small
\setlength{\tabcolsep}{4pt}
\caption{Number of tokens, GFLOPs and Zero-shot ImageNet-1k vs prior CLIP-efficiency methods}
\label{tab:main}
\begin{tabular}{lccccc}
\hline
Method  & Res & Tokens & GFLOPs & IN1k \\
\hline
TinyCLIP (ViT-8M/16)          & 224 & 197 & 1.59 & 41.1\\
Open CLIP (ViT-B/32)              & 224 & 50 & 4.37 & 62.9  \\
\textbf{\methodname{}-low} (ViT-B/16)     & 224 & 71.93 & 6.22 & 62.6 \\
\hline
TinyCLIP (ViT-39M/16)          & 224 & 197 & 7.55 & 63.5\\
MobileCLIP-S2 (Conv)      & 224 & -- & 7.11 & 64.6  \\
\textbf{\methodname{}-mid} (ViT-B/16)     & 224 & 89 & 7.8 & 64.2 \\
\hline
CLIP ViT-B/16              & 224 & 197 & 16.87 & 66.3  \\
MobileCLIP-B (RegNetY-B)      & 224 & -- & 17.02 & 65.3  \\
MobileCLIP2 (RegNetY-B)     & 224 & -- & 17.02 & 66.0  \\
\textbf{\methodname{}-high} (ViT-B/16)     & 224 & 160 & 14.03 & 66.3 \\
\hline
\end{tabular}
\end{table}

\subsection{Effect of KV Caching}
\label{sec:kv}
Fig.~\ref{fig:cached_vs_naive} isolates KV reuse: we compare \textit{cached} progressive inference (our default) to a \textit{naive forward} that re-encodes all tokens at each refinement. For the same accuracy points, naive forward incurs an additional \mbox{+0.56}, \mbox{+1.01}, \mbox{+2.23}, and \mbox{+2.80 GFLOPs} as resolution increases (up to $\sim$20\% overhead at the deepest point), whereas caching amortizes earlier computation. This confirms that cross-level causality plus KV reuse is critical for practical coarse-to-fine inference.

\section{Conclusion}
\label{sec:conclusion}
We presented \methodname{}, an adaptive-resolution alternative to CLIP that (i) replaces patchify with a multi-level wavelet tokenizer, (ii) using a block-causal cross-level attention mask, and (iii) reuses computation via KV caching. This design enables coarse$\rightarrow$fine inference in a \emph{single} ViT image tower, exposing a simple trade of accuracy vs. compute at test time without retraining multiple models or altering the text tower. \methodname{} attains near-baseline zero-shot accuracy on ImageNet-1k while reducing image-tower GFLOPs at matched accuracy (up to $\sim$30\%), yielding practical compute savings and a deployment-friendly model for resource-constrained scenarios as the compute–performance trade-off can be dynamically adjusted at inference.
For future work, \methodname{} could be particularly beneficial for high-resolution images, where computational demands increase substantially. Higher-resolution inputs would allow the use of additional decomposition levels in the wavelet tokenizer, enabling exponentially higher compute savings by utilizing the early exit strategy presented.

\vfill\pagebreak

\bibliographystyle{IEEEbib}
\bibliography{strings,refs}

\end{document}